\documentclass[runningheads]{llncs}
\titlerunning{FedCLF -- Efficient Participant Selection for FL in IoV Networks}
\usepackage[T1]{fontenc}

\usepackage{graphicx}

\usepackage{algorithm}
\usepackage{algpseudocode}
\usepackage{amsmath}
\usepackage{caption}
\usepackage{cite}
\usepackage{subcaption}
\usepackage{hyperref}
\usepackage{url}
\usepackage{float}
\usepackage{array}
\usepackage{enumitem}
\usepackage{lipsum} 

\usepackage{rotating}

\makeatletter
\renewcommand{\fnum@algorithm}{\fname@algorithm}
\makeatother

\begin{document}

\title{FedCLF -- Towards Efficient Participant Selection for Federated Learning in Heterogeneous IoV Networks}

\author{Kasun Eranda Wijethilake\thanks{The corresponding author sincerely acknowledges the generous support of the Government of the Commonwealth of Australia for funding this research work via its International Research Training Program Scholarship (Allocation No. 20235938).}\and Adnan Mahmood \and
Quan Z. Sheng}

\institute{School of Computing, Macquarie University, Sydney, NSW 2109, Australia
\email{kasuneranda.wijethilake@hdr.mq.edu.au},
\email{\{adnan.mahmood,michael.sheng\}@mq.edu.au}}

\maketitle              

\begin{abstract}
Federated Learning (FL) is a distributed machine learning technique that preserves data privacy by sharing only the trained parameters instead of the client data. This makes FL ideal for highly dynamic, heterogeneous, and time-critical applications, in particular, the Internet of Vehicles (IoV) networks. However, FL encounters considerable challenges in such networks owing to the high data and device heterogeneity. To address these challenges, we propose FedCLF, i.e., FL with Calibrated Loss and Feedback control, which introduces calibrated loss as a utility in the participant selection process and a feedback control mechanism to dynamically adjust the sampling frequency of the clients. The envisaged approach (a) enhances the overall model accuracy in case of highly heterogeneous data and (b) optimizes the resource utilization for resource constrained IoV networks, thereby leading to increased efficiency in the FL process. We evaluated FedCLF vis-à-vis baseline models, i.e., FedAvg, Newt, and Oort, using CIFAR-10 dataset with varying data heterogeneity. Our results depict that FedCLF significantly outperforms the baseline models by up to a \textit{16\%} improvement in high data heterogeneity-related scenarios with improved efficiency via reduced sampling frequency.


\end{abstract}

\section{Introduction}\label{Introduction}

Over the past decade or so, there has been a remarkable surge in the adoption of the emerging and promising paradigm of the Internet of Vehicles (IoV) which is, in fact, a significant advancement beyond conventional vehicular ad hoc networks. This paradigm establishes intricate connections not only amongst the vehicles, but also between the vehicles and the roadside infrastructure, vulnerable pedestrians, and the backbone network \cite{FL-IOV-Zhou-IEEE-2023}. By enabling vehicles to interact with various interconnected entities, IoV holds a substantial promise for enhancing road safety via reducing traffic accidents, mitigating traffic congestion, and facilitating the delivery of diverse information-centric services \cite{FL-IOV-Survey-Xu-TST-2022}. A promising intelligent approach in this context is the Federated Learning (FL), i.e., a decentralized machine learning (ML) technique, wherein multiple clients collaborate for resolving the ML-related problems by training a global model locally without sharing their data. This preserves data privacy and reduces latency which are indispensable in IoV networks \cite{FL-IOV-CC-MUN-2024}.

In FL, training occurs at the client level, however, it is impractical to involve all clients in a single training round owing to higher communication overhead, resource utilization, and delays in the training process. Therefore, only a subset of clients is selected for each training round. This is referred to as the participant selection process and it significantly impacts the overall performance of FL. However, the participant selection process faces significant challenges \cite{FL-Survey-Behnaz-ACM-2022}, which are further exacerbated when extending FL to IoV, as these networks encompass vehicles characterized by high mobility, resource limitations, and unstable connectivity, and mandates rapid model convergence \cite{FL-IOV-Wang-WCNCW-2021}. Whilst a number of approaches have been introduced to address these challenges, the field of participant selection in FL within IoV is still in its infancy. In particular, a critical research gap remains in understanding how to improve the overall model accuracy and the efficiency of the FL process in the context of IoV networks.

We, accordingly, envisage FedCLF (FL with Calibrated Loss and Feedback Control), i.e., an efficient FL system specifically designed for optimizing the participant selection process to achieve an overall high model accuracy in highly dynamic, data heterogeneous, and resource constrained IoV networks. It proposes a novel utility metric to achieve an overall high model accuracy while tackling the high data heterogeneity. It further incorporates a new feedback control mechanism to dynamically adjust the sampling frequency of clients, thereby improving the overall resource utilization which, in turn, enhances the efficiency of the FL process without compromising the model performance. We evaluate FedCLF for image classification purpose in IoV scenarios via a simulation platform under different levels of data heterogeneity. Overall, the salient contributions of this paper are as follows:

\begin{itemize}[label=$\bullet$]

\item We propose a new utility metric for participant selection process in FL to achieve better overall model accuracy;

\item We introduce a feedback control mechanism to dynamically adjust the sampling frequency of the clients, thereby improving the efficiency of FL without degrading the overall model accuracy;

\item We evaluate the envisaged method and demonstrate how it outperforms several state-of-the-art methods.

\end{itemize}

The rest of this paper is organized as follows. We first review the state-of-the-art pertinent to FL in the IoV domain in Section \ref{Related Work}. The system architecture of our envisaged approach is illustrated in Section \ref{System Design}. Moreover, we evaluate our approach vis-à-vis several state-of-the-art baseline models in Section \ref{Experimental Setup and Simulation Results}. Finally, we conclude the paper in Section \ref{Conclusion}.

\section{Related Work}\label{Related Work}

In this section, we discuss how research addresses the challenges of participant selection process in FL, particularly in the context of applying FL to the IoV, and how these studies motivated us to develop our new approach FedCLF.

In order to address issues related to data and device heterogeneity in FL, several statistical and system utilities have been introduced to prioritize and select the participants who can contribute more effectively to the learning process. For example, method introduced in \cite{FL-OORT-Lai-USENIX-2021} combines both statistical (e.g., loss) and system utilities  (e.g., training time) to optimize the participant selection process. Addressing trust-related concerns, a hierarchical trust evaluation model that ranks nodes is introduced in \cite{FL-Trust-WANG-AHN-2023}. Introducing fairness into FL training, a framework that incorporates group and individual fairness is proposed by \cite{FL-GIFAR-FL-Yue-INFORMS-2023}. 

In the context of IoV, research addresses critical challenges specific to applying FL. For instance, an FL-based intrusion detection system within a software-defined networking framework, integrating trust metrics to secure IoV networks is proposed in \cite{FL-IoV-IDS-Hbaieb-ACM-2022}. Addressing resource constraints and data heterogeneity in IoV, a content-based participant selection and wireless resource allocation method, that combines the imbalance of dataset, computational resources, and wireless resources is discussed in \cite{FL-IOV-Wang-WCNCW-2021}. In \cite{FL-IOV-Liang-IEETransaction-2022}, a semi-synchronous FL protocol is introduced to tackle the challenges posed by high vehicle mobility and uncertainty. Most of IoV applications are time and safety critical which require a rapid model convergence. To this extend, a novel utility that combines the training weights with dataset sizes to enhance the overall model accuracy swiftly within stringent time constraints is envisaged in \cite{FL-Zhao-IEEE-2023}. Moreover, they have introduced a feedback control mechanism that enables dynamic adjustments within FL systems crucial for the dynamic IoV network. They discuss utilizing variables such as loss, weight changes, accuracy, client status, and data heterogeneity levels as feedback signals to influence FL configurations such as sampling frequency, participant selection process, learning rate, and threshold times, thereby adapting the FL process effectively.

Our work is motivated by this concept that FL with a feedback control mechanism provides an excellent answer for the extremely dynamic IoV network. We applied this concept to the sampling frequency adjustment depending on the performance of the overall model accuracy, along with a new statistical utility measurement for participant selection process which will be discussed in the rest of this paper.

\section{System Design}\label{System Design}

In this section, we introduce our novel system FedCLF comprising two integral components: the introduction of a novel utility measurement and the integration of a feedback control mechanism for determining the sampling frequency. Subsequent sections delve into a comprehensive discussion of these two pioneering approaches, explaining their methodologies and implications in greater detail. 

\subsection{System Architecture}\label{System Architecture}

\begin{figure}[t]
    \centering
    \includegraphics[width=0.75\linewidth]{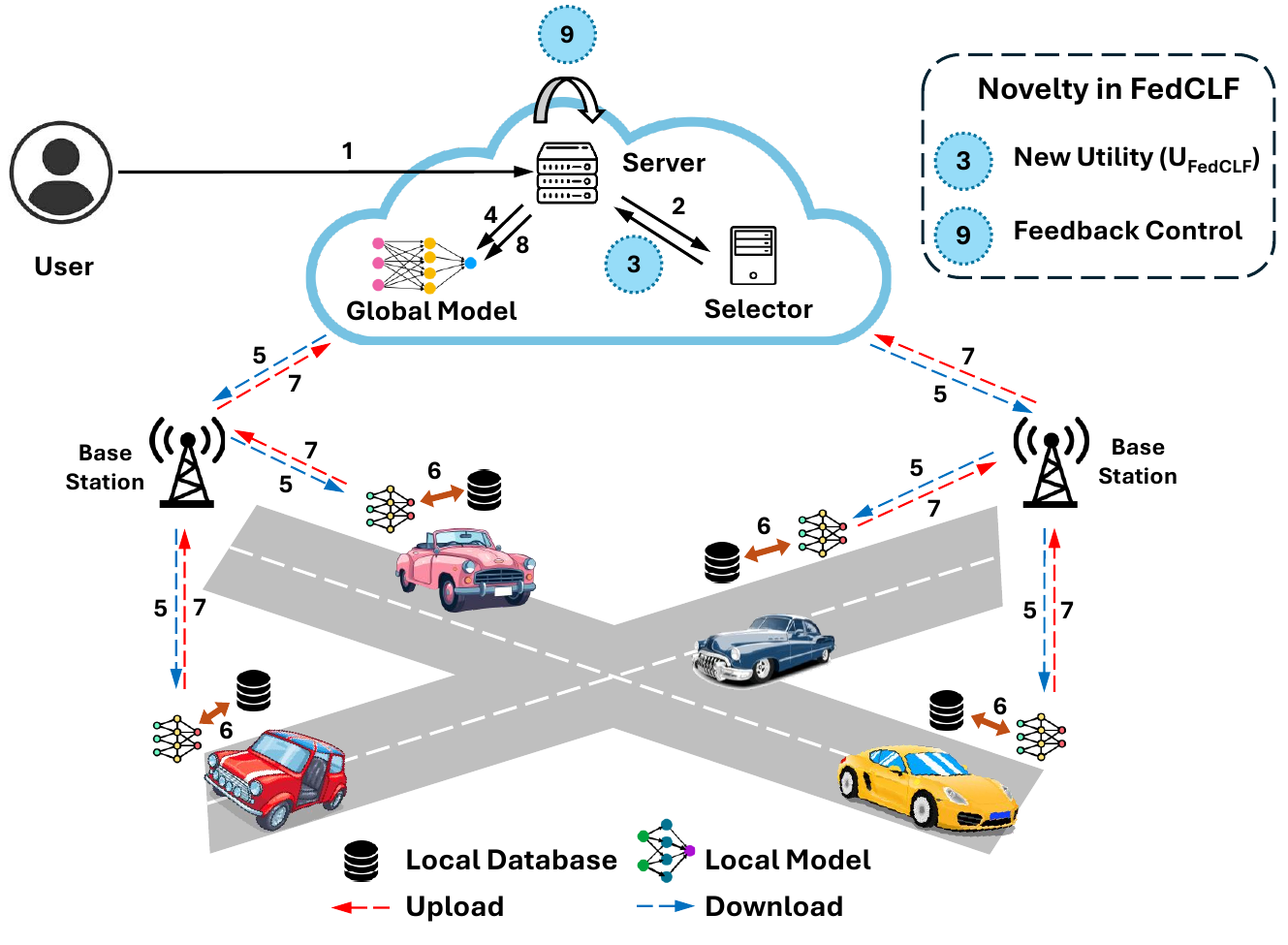}
    \caption{FedCLF architecture in the context of IoV networks.}
    \label{fig:system architecture}
\end{figure}

Fig. \ref{fig:system architecture} illustrates the envisaged FL architecture in the context of an IoV network \cite{FL-IOV-Zhou-IEEE-2023}. It deviates from the typical FL scenario in steps \textit{3} and \textit{9} since it introduces (a) a new utility for improving the overall model accuracy and (b) a feedback control component for optimizing the resource utilization while maintaining the model accuracy. Each step in the architecture is sequentially numbered and the associated task vis-\`a-vis each step is described as follows.

\textbf{\textit{Step 1 --}} The service requester in an IoV network initiates the FL request to the server. This request includes all indispensable parameters for the training process, including but not limited to the number of participants, the number of local epochs, the number of training rounds, learning rate, aggregation technique, sampling method, and training time per round.

\textbf{\textit{Step 2 --}} The server forwards the request to the selector tasked with selecting the requisite number of clients for training from the available pool of clients. In an IoV network, the clients encompass numerous entities, i.e., vehicles, roadside units (RSUs), and mobile edge devices. 

\textbf{\textit{Step 3 --}}  The selector chooses the necessary number of clients from the available pool based on the utility considered in the FL training process and relays the selected client details back to the server. In our envisaged system, calibrated loss serves as the new utility (\(U_{FedCLF}\)). As only selected clients undergo training, the loss of all clients is not calculated at each round. To address this, our system calibrates the loss of other clients not chosen in the last training round, factoring in the change of loss in the global model. This process is elaborated in Section \ref{Model Client Selection Utility}. 

\textbf{\textit{Step 4 --}} The server initializes the global model randomly. This step takes place at the first cycle of the FL process only as in the other cycles, the initiated global model will be updated based on client updates.

\textbf{\textit{Step 5 --}} The server passes the parameters of the global model to the selected clients for training with all required other parameters of the FL process. 

\textbf{\textit{Step 6 --}} Then the clients train the global model received from the server based on the available local data at their disposal and the training parameters received from the server. In the real world scenarios, due to the heterogeneity of the data in the IoV network, data available with them will be non independent and identically distributed (Non-IID), and different in sizes as well.  

\textbf{\textit{Step 7 --}} In this step, clients upload the trained model parameters back to the server. 

\textbf{\textit{Step 8 --}} The server aggregates all received model updates from the selected clients and updates its global model. In our envisaged architecture, weighted averaging (as in FedAvg \cite{FL-FedAvg-McMahan-AISTATS-2017}) is employed as the aggregation technique. 

\textbf{\textit{Step 9 --}} This is the feedback control step introduced in our envisaged system. This step begins from the third round onwards, as it compares the overall model accuracy from the last round with that of the previous round. In our approach, participant selection process (steps \textit{2} and \textit{3}) will occur only if the overall model accuracy at the last round is lower than that of the previous round. Further details are provided in Section \ref{Model-Feedback}.

The FedCLF methodology is delineated in the Algorithm. Overall, the FL process is controlled by the $ServerUpdate$ procedure which selects clients via the $Selection$ procedure. Subsequently, the chosen clients undergo training via the $ClientUpdate$ procedure. The feedback control mechanism takes place at the $ServerUdpate$. It monitors the fluctuations in the overall model accuracy at the end of each round and dynamically regulates the sampling frequency as delineated from lines \textit{4} to \textit{8} of the Algorithm. Additionally, the calculation of our new utility \(U_{FedCLF}\) is depicted in line \textit{24}. These fundamental aspects of the FedCLF methodology lay the groundwork for further exploration in subsequent sections of this paper. 

\begin{algorithm}[]
\caption{Pseudocode of the FedCLF framework}\label{alg:cap}
\label{alg:Pseudo-Code}
\begin{algorithmic}[1]
\State Parameters: Round Number - $r$, Total Number of Rounds - $R$, Number of Selected Clients - $k$, Selected Clients at Round $r$ - $C_r$, Total Number of Clients - $K$, Initial Model Weights - $w_0$, Number of Samples - $n$, Number of Epochs - $E$, Learning Rate - $\eta$, Accuracy at Round $r$ - $Acc_r$, Loss of Client $k$ - $l_k$, Local Data - $d$, Cost Function to Calculate Loss on Data $d$ - $J(w,d)$

\Procedure{ServerUpdate}{$k, w$}
    \For{$r = 1, 2, ..., R$}
        \If{$Acc_{r-1} < Acc_{r-2}$} \Comment{Feedback Control}
            \State $C_r \gets Selection(k, Acc_{r-1}, Acc_{r-2})$
        \Else
            \State $C_r \gets C_{r-1}$ 
        \EndIf

        \For{$c_k \in C_r$ in parallel} 
            \State $w_r^k, l_r^k, n^k \gets ClientUpdate(k,w_{r-1})$ 
        \EndFor
        \State $w_r \gets \sum_{k} \frac{n_k}{n} w_r^k$
        
    \EndFor
\EndProcedure

\Procedure{Selection}{$k, Acc_{r-1}, Acc_{r-2}$}
    \If{$r \leq K/k$}\Comment{Unique Sampling}
        \State $C_{Available} \gets \{c \in C_K \,|\, c \notin C_{Sampled}\}$
        \State $C_r \gets rand(C_{available},k)$
        \State $C_{Last Round} \gets C_r $
        \State $C_{Sampled} \gets C_{Sampled} \cup C_r $
    \Else
        \For{$c_k \in C_K$} 
            \If{$C_k \notin C_{Last Round}$}
                \State $U_{FedCLF}^k \gets l_k\frac{Acc_{r-1}}{Acc_{r-2}}$\Comment{Calibrated Loss as \(U_{FedCLF}\)}
            \Else
                \State $U_{FedCLF}^k \gets l_k$
            \EndIf   
        \EndFor
        \State $C_r \gets$ Top $k$ clients that have maximum \(U_{FedCLF}\)
    \EndIf  
    \State $Return$ $C_r$
\EndProcedure

\Procedure{ClientUpdate}{$k, w_r$}
    \For{$e = 1, 2, ..., E$}
        \State $n_k \gets$ Number of samples at the client
        \State $l_k \gets J(w_r, d_k)$
        \State $w_r^k \gets w_r - \eta \nabla J(w_r, d_k)$
    \EndFor
    \State $Return$ $w_r^k, l_k, n_k$ 
\EndProcedure

\end{algorithmic}
\end{algorithm}

\subsection{Participant Selection Utility}\label{Model Client Selection Utility}

In FL, achieving time to accuracy performance relies on two salient metrics, i.e., statistical efficiency and system efficiency, as determined by the number of rounds required to attain a predefined target accuracy and the duration of each round, respectively \cite{FL-OORT-Lai-USENIX-2021}. Statistical efficiency primarily relies on the data stored on the client thereby indicating its capability to contribute meaningfully to the learning process. Conversely, system efficiency pertains to the speed at which clients can execute their tasks for influencing the overall pace of the FL operations.

In order to achieve a higher statistical efficiency, it is crucial to assign a higher utility value to clients whose data can significantly influence the model divergence. The optimal solution for this is outlined in Equation \ref{eq1}, wherein clients exhibiting a larger aggregate gradient norm across all their samples are prioritized \cite{FL-OORT-Lai-USENIX-2021} \cite{FL-Survey-Fu-IEEE-2023}:

\begin{equation}\label{eq1}
U=|B_i|\sqrt{\frac{1}{|B_i|}\sum_{k \in B_i}||\nabla f(k)||_2^2}
\end{equation}
here, \(||\nabla f(k)||_2\) implies the \(L_2\)-norm of each data sample \textit{k}'s gradient within the (data) bin \textit{B} of the FL client \textit{i}. Employing the aforementioned calculation method for utility entails increased computational overhead resulting in higher delays in the FL process. As an alternative, loss can be utilized since it often correlates with the gradient norm and is already available during the local training. Equation \ref{eq2} illustrates this approach \cite{FL-Survey-Fu-IEEE-2023}:

\begin{equation}\label{eq2}
U=|B_i|\sqrt{\frac{1}{|B_i|}\sum_{k \in B_i}Loss(k)^2}
\end{equation}

In the literature, loss is often considered directly as the utility or with a combination of another attribute like sample size as the utility \cite{FL-OORT-Lai-USENIX-2021}. However, a limitation arises from the inability to obtain current loss values for all clients after each training round. This constraint arises as loss is calculated only for the selected clients during the training process. Consequently, clients not selected for training will have loss values calculated from a previous round thereby potentially resulting in higher deviations compared to actual loss values. This discrepancy can lead to incorrect participant selection in subsequent rounds. 

To address the above limitation, in our approach, we prioritize the statistical efficiency by employing the calibrated loss value as the new utility (\(U_{FedCLF}\)) metric within the FedCLF framework. This strategic choice optimizes the learning process by leveraging client data to achieve desired accuracy levels in minimal rounds. This focus on statistical efficiency is especially crucial in heterogeneous data environments like the IoV network, where dynamic data distribution presents challenges. By effectively harnessing client data despite its diversity, our approach meets the demands of dynamic environments, ensuring effective learning outcomes amidst fluctuating data conditions. \(U_{FedCLF}\) is calculated as in Equation \ref{eq3}:

\begin{equation}\label{eq3}
U_{FedCLF} = \begin{cases} 
      |B_i|\sqrt{\frac{1}{|B_i|}\sum_{k \in B_i}Loss(k)^2} & \text{: for } c \in C_{Selected} \\
      (|B_i|\sqrt{\frac{1}{|B_i|}\sum_{k \in B_i}Loss(k)^2}) (\frac{Loss_{r-1}}{Loss_{r-2}}) & \text{: for } c \notin C_{Selected}
   \end{cases}
\end{equation}
where \(C_{Selected}\) refers to the clients sampled at the last training round (\textit{r-1}). For clients who trained at the last round, \(U_{FedCLF}\) will be the loss as it is whereas for other clients, \(U_{FedCLF}\) is the calibrated loss by adding a correction factor to previously calculated loss values. This correction factor is the ratio of the loss value of the global model calculated at the last two consecutive rounds (\textit{r-1} and \textit{r-2} rounds), computed as \(\frac{Loss_{r-1}}{Loss_{r-2}}\). The enhancement in the overall model accuracy resulting from the incorporation of this correction factor into \(U_{FedCLF}\) instead of utilizing the loss directly as the utility is elaborated further through simulation results in Section \ref{Experimental Setup and Simulation Results}.

\subsection{Feedback Control}\label{Model-Feedback}

In our envisaged system, we introduce a crucial feature known as the feedback control mechanism which adds a dynamic dimension to the FL process. This mechanism allows for the adaptation of the FL process in response to changes in the FL environment, thereby avoiding rigidity to the initially configured FL process. This capability is particularly advantageous in highly dynamic environments such as IoV, where conditions can vary rapidly, enabling more flexible and efficient FL operations.

In \cite{FL-Zhao-IEEE-2023}, when discussing the setting of sampling frequency considering the levels of data heterogeneity among clients, they emphasize that there will be either no decline or only a slight decline in overall model accuracy performance. In such a scenario, the sole advantage of adjusting the sampling frequency would be the minimization of resource utilization, as there would be no need for sampling at each round. However, in contrast to this perspective, our envisaged method demonstrates an improvement in the overall model accuracy by dynamically changing the sampling frequency based on our feedback control mechanism. This improvement is further elaborated through simulation results in Section \ref{Evaluations-Feedback}. In our system, we utilize the overall model accuracy at the conclusion of each round as the feedback to determine the sampling frequency. The rationale behind this approach is straightforward: if the overall model accuracy in the FL process demonstrates improvement or stability, we maintain the same selected client set for subsequent rounds. Conversely, if the overall model accuracy shows a decline, we opt for a different set of clients for the next round. This adaptive strategy allows us to dynamically adjust the client sampling process based on the observed performance of the global model, thereby enhancing the efficiency and effectiveness of FL. 

In this scenario, numerous avenues for research are opened. Instead of solely monitoring the decline in overall model accuracy, alternative approaches can be explored. For instance, researchers can consider setting static or dynamic threshold levels for the decline in accuracy. Additionally, predictive models can be developed to anticipate the next expected accuracy based on historical trends and by comparing this predicted accuracy with the actual observed accuracy, suitable actions can be implemented real time.

\section{Experimental Setup and Simulation Results}\label{Experimental Setup and Simulation Results}

In our simulations, we employed the Flower framework \cite{FL-Flower-Beutel-2020} to implement and simulate the envisaged method across diverse datasets. Flower, an open-source platform licensed under Apache \textit{2.0}, has gained widespread adoption by researchers in both academia and industry. We selected Flower for its scalability, flexibility, and user-friendly interface which facilitated the development of our FL system. For training of the model, a basic convolutional neural network (CNN), i.e., commonly used for image classification tasks, was used. It is a simple feed-forward CNN with two convolutional layers followed by max pooling layers and three fully connected layers. 

For purpose of simulations, we maintained consistent base configurations: total number of participants -- \textit{50}, number of selected clients per round -- \textit{5}, total number of rounds -- \textit{100}, number of local epochs -- \textit{1}, and learning rate -- \textit{0.001}. We employed a moving average with a window size ($N$) of \textit{30} as depicted in Equation \ref{Eq-MA} to plot the accuracy in a bid to ensure a smooth curve:

\begin{equation}\label{Eq-MA}
    MA_r = \frac{1}{N} \sum_{i=0}^{N-1} Ac_{r-i}
\end{equation}
here, \(MA_r\) implies the moving accuracy at the round number r and \(Ac_{r-i}\) denotes the accuracy at the round number \(r-i\). Furthermore, we compared the performance of our envisaged method with several baseline models, FedAvg (which employs random participant selection process), Oort \cite{FL-OORT-Lai-USENIX-2021} (which utilizes a combination of loss and training time as a utility), and Newt \cite{FL-Zhao-IEEE-2023} (which factors in weight change during training and the amount of local client data as a utility). These comparisons provided valuable insights into the effectiveness of our approach relative to the existing baseline models.

In the forthcoming sections, we will delve into the simulation results covering four key areas: preparation of heterogeneity datasets, new utility metric (\(U_{FedCLF}\)), feedback control mechanism and overall evaluation of FedCLF.

\subsection{Preparation of Heterogeneous Datasets}\label{Preparation of Heterogeneous Datasets}

To assess the efficacy of the envisaged system, it is indispensable that the datasets employed for the simulation purposes closely resemble the real-world scenarios. Since the key intent of the envisaged model pertains to an IoV network, datasets should exhibit significant heterogeneity. Additionally, by varying the heterogeneity levels within the datasets and for conducting simulations across these variations, we can realistically evaluate a system's performance under diverse conditions. This approach enables us to observe how the system responds and adapts to different scenarios, thereby providing valuable insights into its robustness and effectiveness.

For the sake of our simulations, we utilized the CIFAR-10 dataset \cite{FL-CIFAR10-Krizhevsky-2009}, encompassing \textit{60,000} colored images (\textit{32} x \textit{32} resolution) categorized into \textit{10} classes each with \textit{6,000} images. Out of the \textit{60,000} images, \textit{50,000} were employed for training and \textit{10,000} for testing purposes. In FL, data heterogeneity entails two key aspects, i.e., data label distribution and data size. To ensure datasets with varying levels of data heterogeneity, we adopted a data partitioning method that involves sorting the data based on class labels and then splitting them into groups of size \textit{S} \cite{FL-FedAvg-McMahan-AISTATS-2017}. Subsequently, these groups are shuffled and combined again to avoid having groups with the same types of labels closer to each other. Finally, based on the splitting method, i.e., equal or non-equal, the arranged dataset is split among the total number of participants. By adjusting the value of \textit{S}, we can manipulate the heterogeneity level of the dataset. To assess the variance in heterogeneity levels across the developed datasets via the above method, we have employed the Earth Mover's Distance (EMD) metric \cite{FL-Zhao-IEEE-2023}. This metric compares the developed dataset with its IID dataset by highlighting the disparity between the two distributions, which is mathematically expressed in Equation \ref{Eq-EMD}:

\begin{figure}[!t]
    \centering
    \begin{subfigure}[b]{0.47\textwidth}
         \centering
         \includegraphics[width=\textwidth]{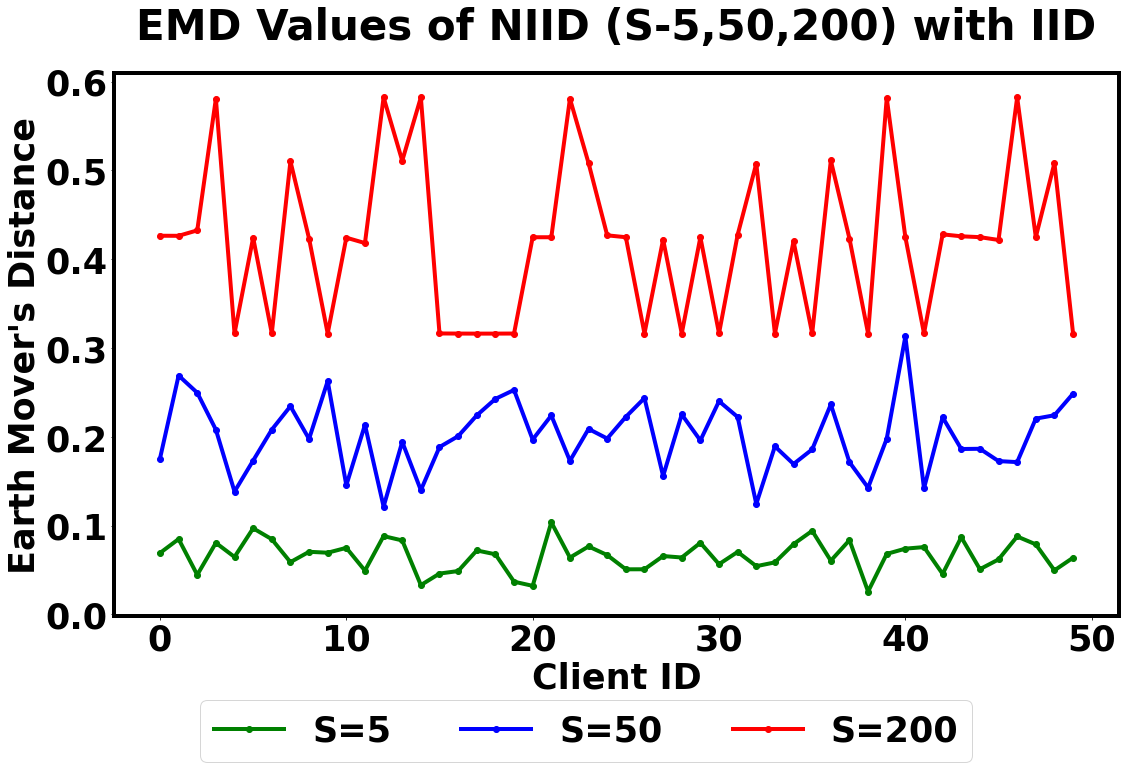}
         \caption{}
         \label{fig:EMDa}
    \end{subfigure}
     \hfill
     \begin{subfigure}[b]{0.47\textwidth}
         \centering
         \includegraphics[width=\textwidth]{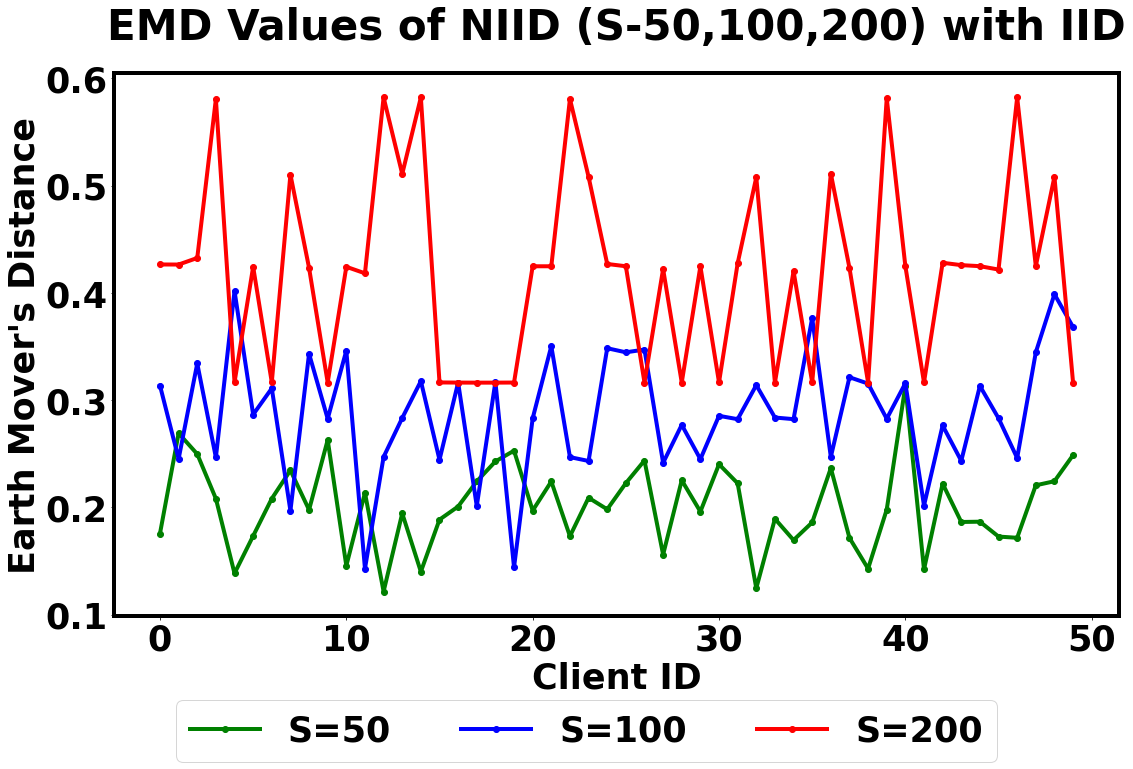}
         \caption{}
         \label{fig:EMDb}
     \end{subfigure}
     \hfill
        \caption{Variance in heterogeneity levels across the developed datasets based on different \textit{S} values.}
        \label{fig:EMD}
\end{figure}

\begin{table}[!b]
\centering
\caption{Datasets with different levels of data heterogeneity.}

\begin{tabular}{ >{\centering\arraybackslash}m{2.5 cm}|>{\centering\arraybackslash}m{3.5cm}|>{\centering\arraybackslash}m{2.5cm}|>{\centering\arraybackslash}m{2.5cm} } 
 \hline
 \textbf{Dataset} & \textbf{Heterogeneity Type} & \textbf{Average EMD} & \textbf{Split Method} \\ 
 \hline
 IID-E & IID & NA & Equal\\ 
 \hline
 IID-NE & IID & 0.03 & Non-Equal\\ 
 \hline
 NIID-S5-E & Non IID (S-5) & 0.07 & Equal\\ 
 \hline
 NIID-S5-NE & Non IID (S-5) & 0.07 & Non-Equal\\ 
 \hline
 NIID-S50-E & Non IID (S-50) & 0.20 & Equal\\ 
 \hline
 NIID-S50-NE & Non IID (S-50) & 0.19 & Non-Equal\\ 
 \hline
 NIID-S200-E & Non IID (S-200) & 0.42 & Equal\\ 
 \hline
 NIID-S200-NE & Non IID (S-200) & 0.40 & Non-Equal \\
 \hline
\end{tabular}

\label{tab:datasets}
\end{table}

\begin{equation}
\label{Eq-EMD}
    EMD=\sum_{i=1}^{C}||P_a(y=i)-P_b(y=i)||
\end{equation}
where \textit{C} represents the total number of classes, and \(P_a\) and \(P_b\) represent the distributions of the developed dataset and its IID dataset, respectively. In our simulations, with \textit{50} clients participating, we segmented the dataset into \textit{50} data segments. To visualize the distinct levels of heterogeneity of the developed datasets, we plotted the EMD values for various \textit{S} values as depicted in Fig. \ref{fig:EMD}. It can be observed that by increasing \textit{S}, the degree of heterogeneity increases. 

Table \ref{tab:datasets} depicts the eight developed datasets to demonstrate different levels of data heterogeneity. \textit{S} values of \textit{5}, \textit{50}, and \textit{200} were employed to ensure a clear variance of EMDs amongst the datasets as portrayed in Fig \ref{fig:EMDa}. When splitting the dataset equally, the same was divided uniformly among \textit{50} clients with each client having \textit{1,000} training samples. For non-equal splitting, we employed a random method to allocate a random number of samples to each client.

\subsection{Evaluation of the New Utility Metric (\(U_{FedCLF}\))}\label{Evaluation of New Utility Metric}

\begin{figure}[!t]
    \centering
    \begin{subfigure}[b]{0.47\textwidth}
         \centering
         \includegraphics[width=\textwidth]{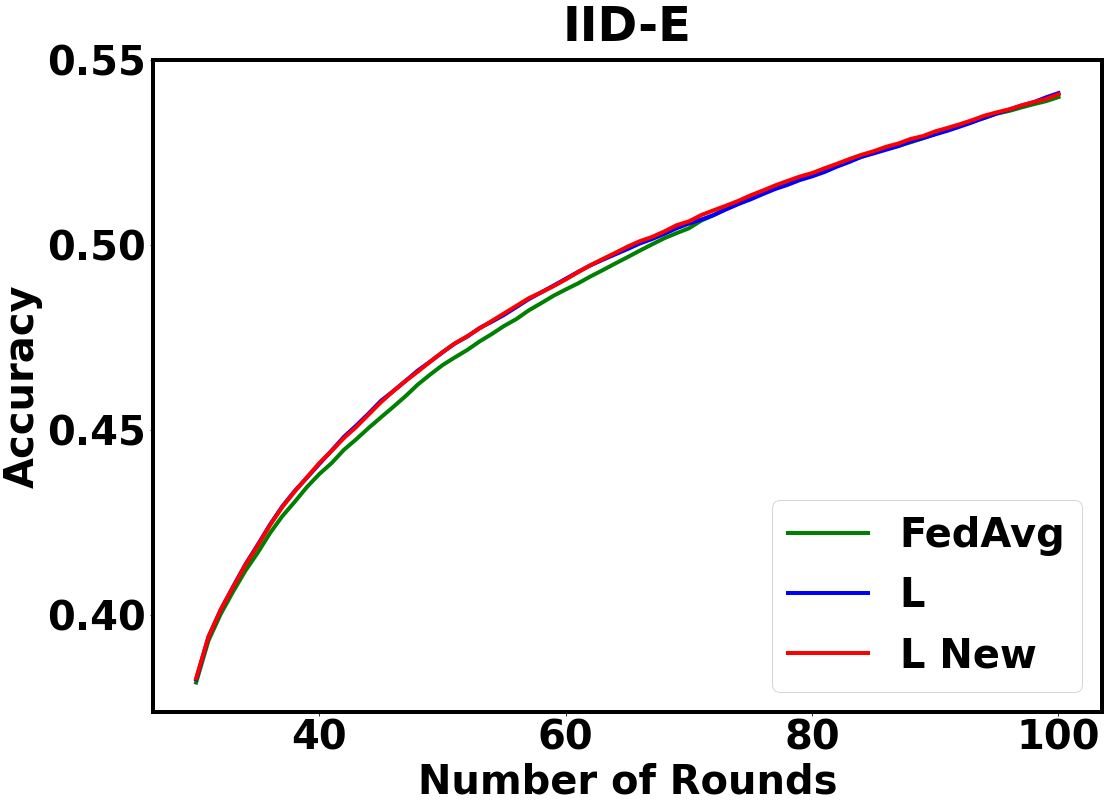}
         \caption{}
         \label{fig:a}
    \end{subfigure}
     \hfill
     \begin{subfigure}[b]{0.47\textwidth}
         \centering
         \includegraphics[width=\textwidth]{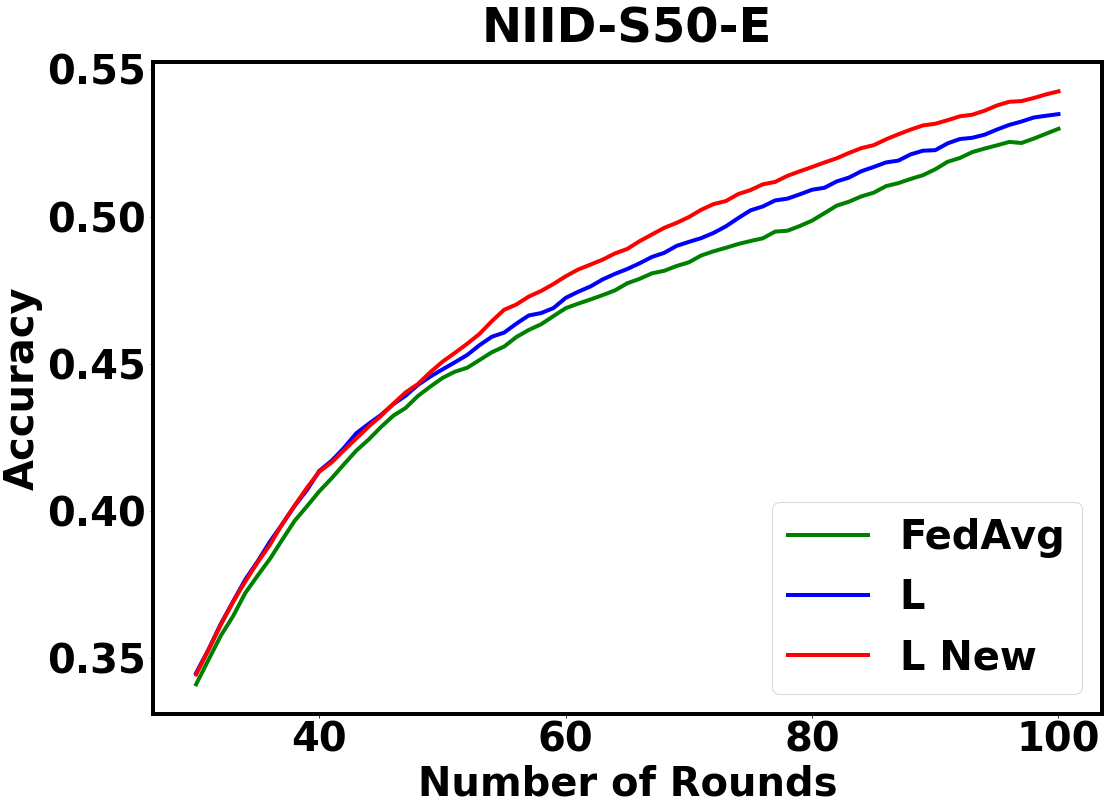}
         \caption{}
         \label{fig:b}
     \end{subfigure}
     \hfill
        \caption{Comparison of the moving average accuracy when using \(U_{FedCLF}\) (\textit{L New}) vis-à-vis \(U_{Loss}\) (\textit{L}) and FedAvg on datasets, IID-E and NIID-S50-E.}
        \label{fig:Plots of U FedCLF}
\end{figure}

In FedCLF, the newly introduced utility metric (\(U_{FedCLF}\)) is the calibrated loss. This section demonstrates the superior performance of \(U_{FedCLF}\) in contrast to the non-calibrated loss as the utility metric (\(U_{Loss}\)). Additionally, the results are compared with the FedAvg, which is one of the most commonly used baseline models in FL (comprehensive analysis is in Section \ref{Overall Evaluation of FedCLF}). Simulations were conducted using the IID (IID-E) and NIID (NIID-S50-E) datasets, and the results are illustrated in Fig. \ref{fig:Plots of U FedCLF}, wherein $L$ and \textit{L New} refer to the simulation results when using \(U_{Loss}\) and \(U_{FedCLF}\) as utilities, respectively. As evident, the results for \textit{L New}, \textit{L}, and the FedAvg are similar in the IID scenario (IID-E). However, a clear distinction emerged as the degree of data heterogeneity increases (NIID-S50-E), i.e., when using \(U_{FedCLF}\), there is a visible improvement of approximately \textit{4\%} over FedAvg and \textit{2\%} over \(U_{Loss}\).

\subsection{Evaluation of Feedback Control}\label{Evaluations-Feedback}

\begin{figure}[!t]
    \centering
    \begin{subfigure}[b]{0.47\textwidth}
        \centering
        \includegraphics[width=\textwidth]{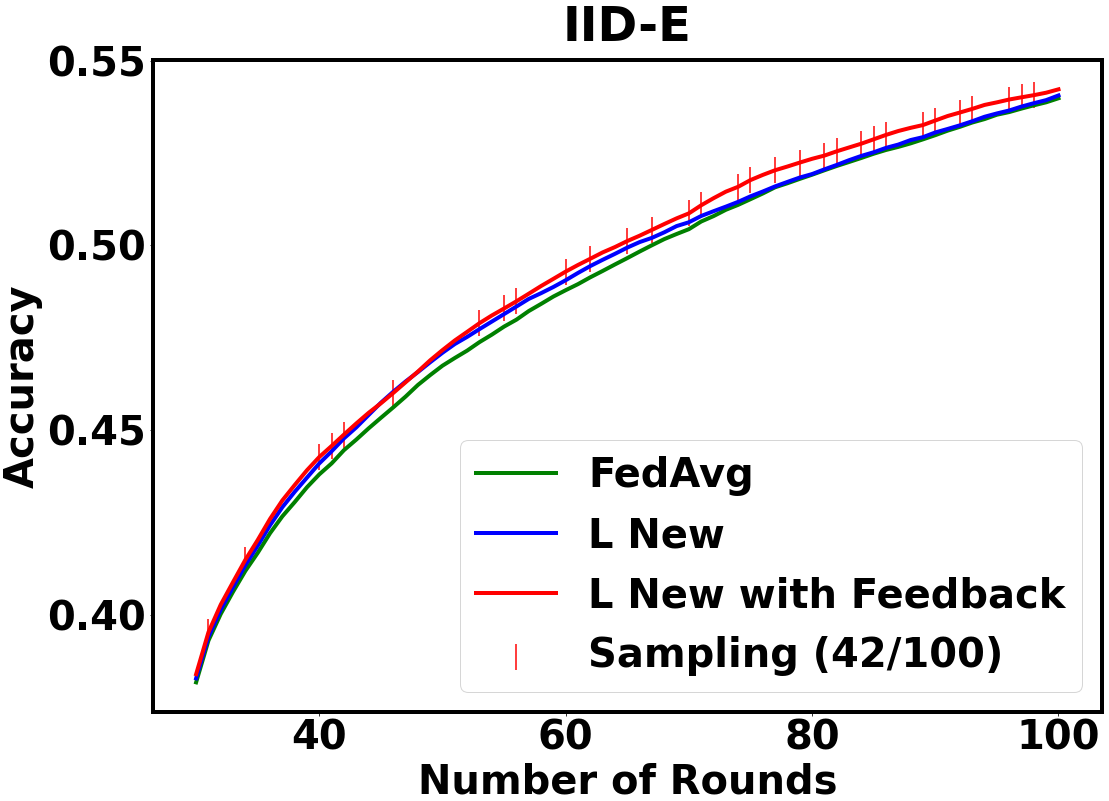}
        \caption{}
        \label{fig:a}
    \end{subfigure}
    \hfill
    \begin{subfigure}[b]{0.47\textwidth}
        \centering
        \includegraphics[width=\textwidth]{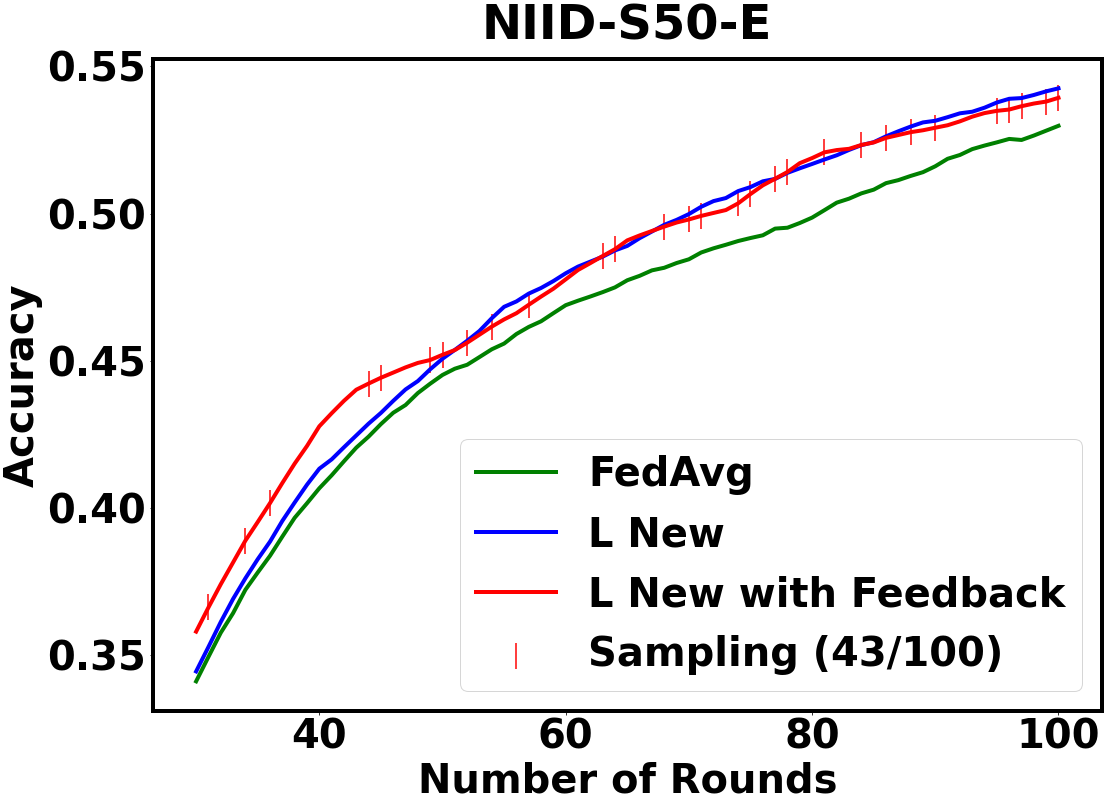}
        \caption{}
        \label{fig:b}
    \end{subfigure}
    \caption{Comparison of the moving average accuracy when using \(U_{FedCLF}\) with a feedback control (\textit{L New with Feedback}) vis-à-vis \textit{L New} and FedAvg on datasets, IID-E and NIID-S50-E.}
    \label{fig:Plots of feedback}
\end{figure}

In this particular section, we discuss how the variation in the sampling frequency based on the newly introduced feedback control mechanism in our envisaged approach performs better than maintaining a constant sampling frequency (as in the traditional FL process). Simulations were conducted using the IID (IID-E) and NIID (NIID-S50-E) datasets, and the results were compared with FedAvg, as depicted in Fig. \ref{fig:Plots of feedback}, wherein \textit{L New} refers to the use of \(U_{FedCLF}\) with sampling at each round and \textit{L New with Feedback} implies to the use of \(U_{FedCLF}\) with the varying sampling frequency based on the feedback control. To illustrate the impact of the feedback control on the sampling frequency, specific sampled points are marked on the \textit{L New with Feedback} subplot with a legend record as $Sampling$. This legend record suggests details on the total number of sampling occasions out of the total number of training rounds. For instance, in the IID-E scenario, sampling occurred in \textit{42} out of \textit{100} rounds, whereas in the case of NIID-S50-E scenario, sampling occurred in \textit{43} rounds. 

A clear improvement to the tune of \textit{1\%} can be observed for \textit{L New with feedback} vis-à-vis both FedAvg and \textit{L New} even in the case of the IID (IID-E) dataset. For the NIID (NIID-S50-E) dataset, \textit{L New with feedback} initially outperforms \textit{L New} with an improvement of approximately \textit{4\%} before converging to similar performance towards the end. Additionally, it reflects an improvement of around 5\% over FedAvg. As discussed in Section \ref{Model-Feedback}, these results demonstrate that by adjusting the sampling frequency based on system performance, we can achieve improved or comparable performance to what is attained when sampling occurs at each round. The significant benefit, however, lies in the reduced resource utilization due to the decreased sampling frequency which is particularly advantageous in resource-constrained networks such as the IoV. Moreover, the plots reveal that the sampling rate is low in the initial rounds and increases as the number of rounds progresses (the concentration of sampling points increases with the number of rounds). This indicates that the efficiency of the overall FL process is higher in the initial rounds, allowing for the possibility of achieving an initial target accuracy within a short period of time. This is a crucial advantage in IoV where most applications are safety and time critical.

\subsection{Overall Evaluation of FedCLF}\label{Overall Evaluation of FedCLF}

\begin{figure}[!htbp]
    \centering
    \begin{subfigure}[b]{0.49\textwidth}
         \centering
         \includegraphics[width=\textwidth]{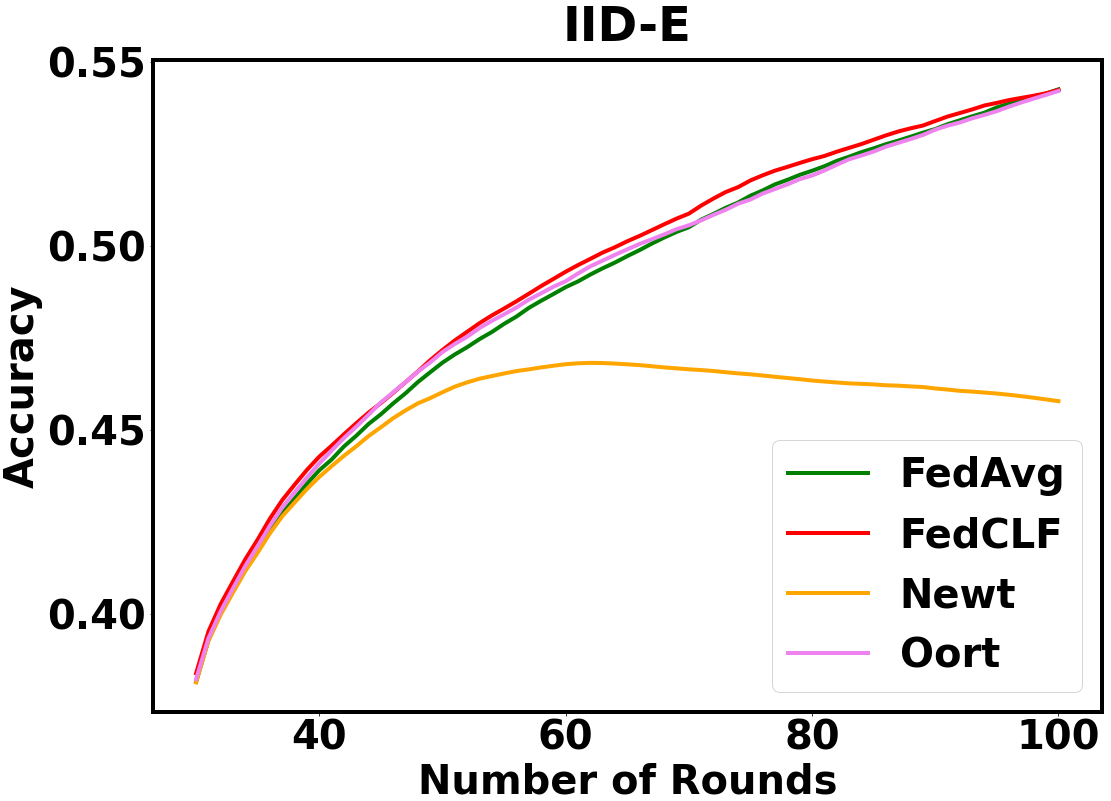}
         \caption{}
         \label{fig:a}
    \end{subfigure}
    \hfill
    \begin{subfigure}[b]{0.49\textwidth}
         \centering
         \includegraphics[width=\textwidth]{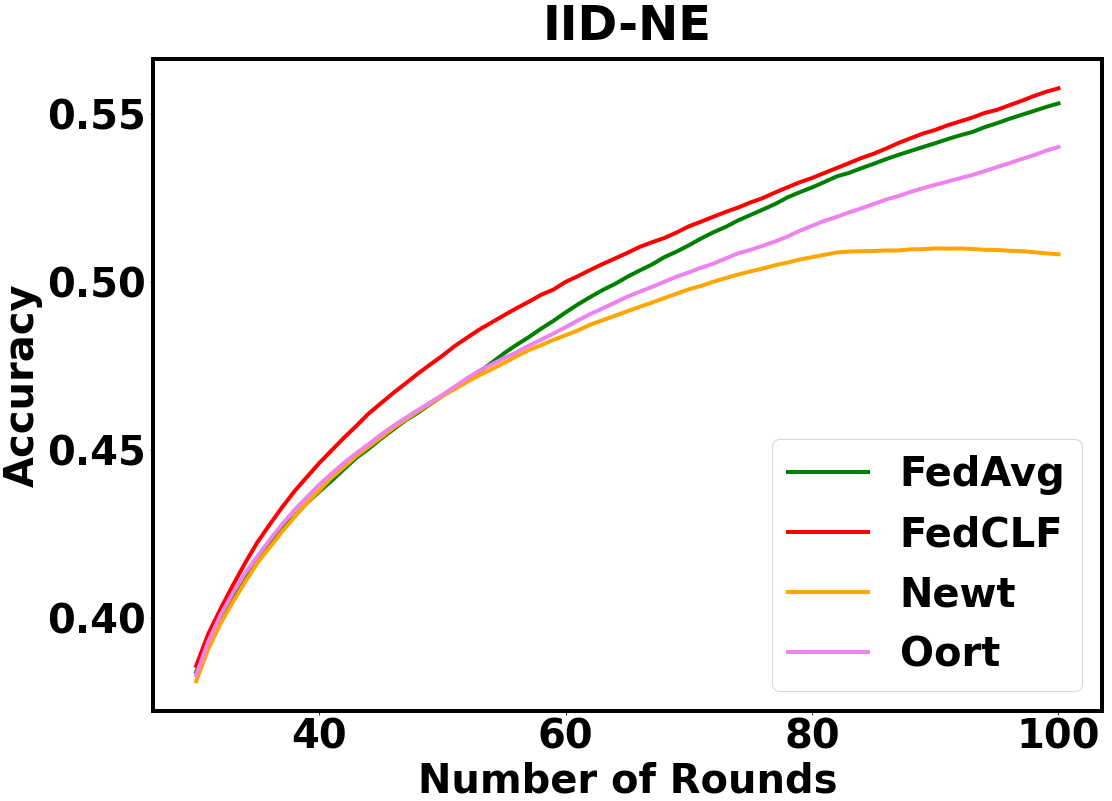}
         \caption{}
         \label{fig:b}
    \end{subfigure}
    \hfill
    \begin{subfigure}[b]{0.49\textwidth}
         \centering
         \includegraphics[width=\textwidth]{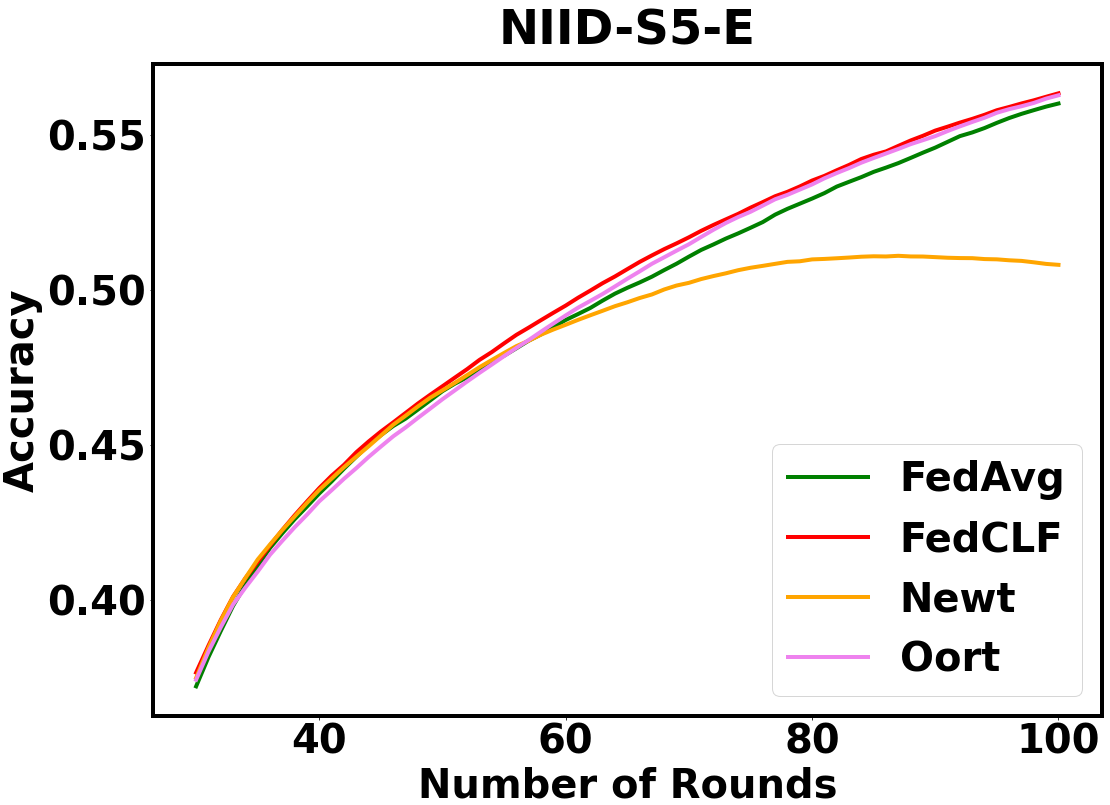}
         \caption{}
         \label{fig:c}
    \end{subfigure}
    \hfill
    \begin{subfigure}[b]{0.49\textwidth}
         \centering
         \includegraphics[width=\textwidth]{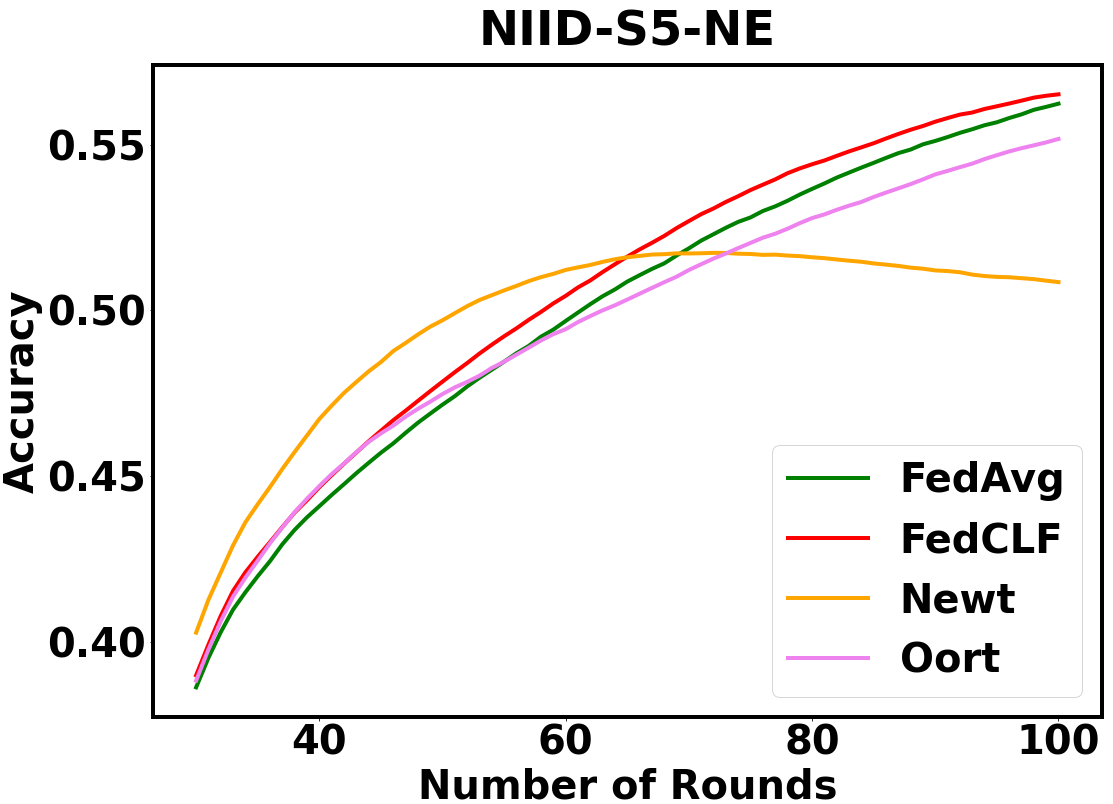}
         \caption{}
         \label{fig:d}
    \end{subfigure}
    \hfill
    \begin{subfigure}[b]{0.49\textwidth}
         \centering
         \includegraphics[width=\textwidth]{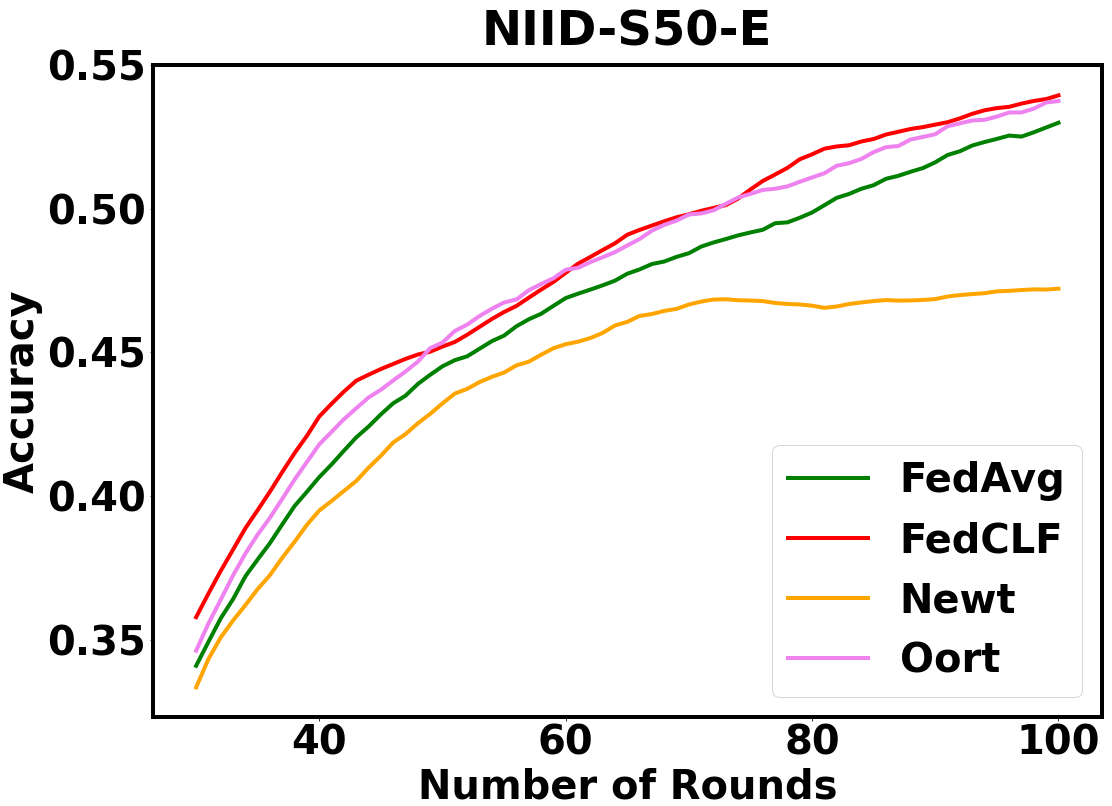}
         \caption{}
         \label{fig:e}
    \end{subfigure}
    \hfill
    \begin{subfigure}[b]{0.49\textwidth}
         \centering
         \includegraphics[width=\textwidth]{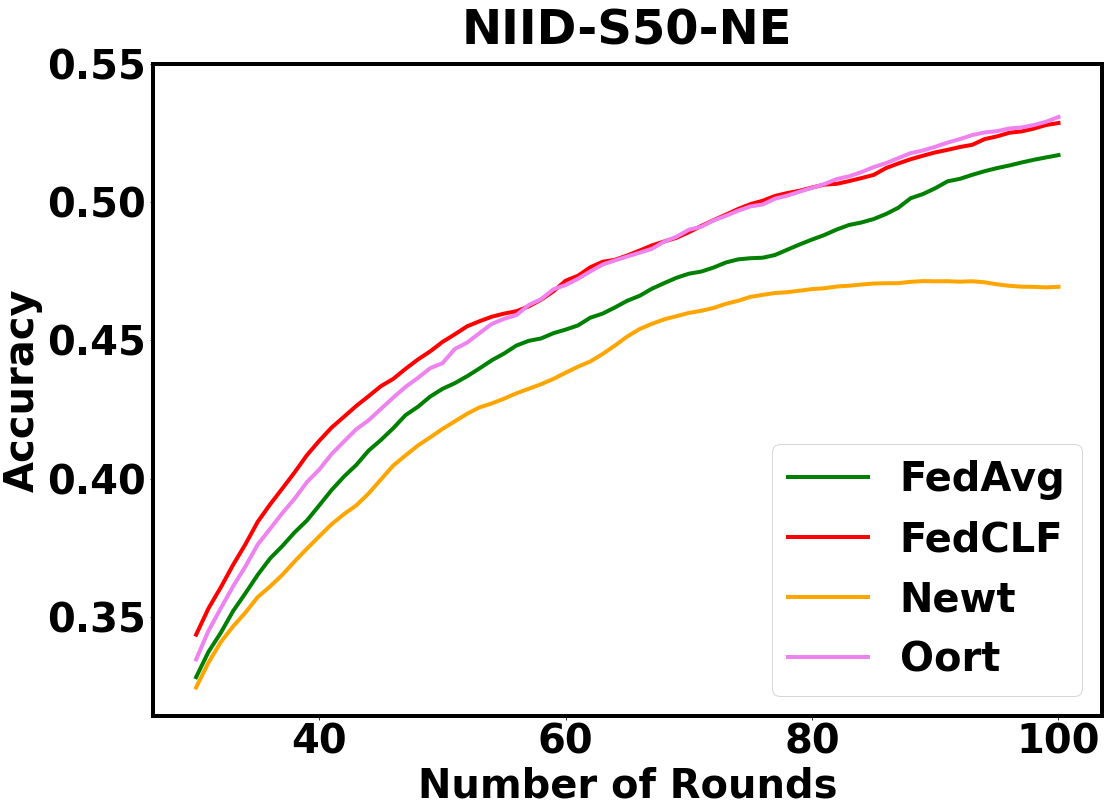}
         \caption{}
         \label{fig:f}
    \end{subfigure}
    \hfill
    \begin{subfigure}[b]{0.49\textwidth}
         \centering
         \includegraphics[width=\textwidth]{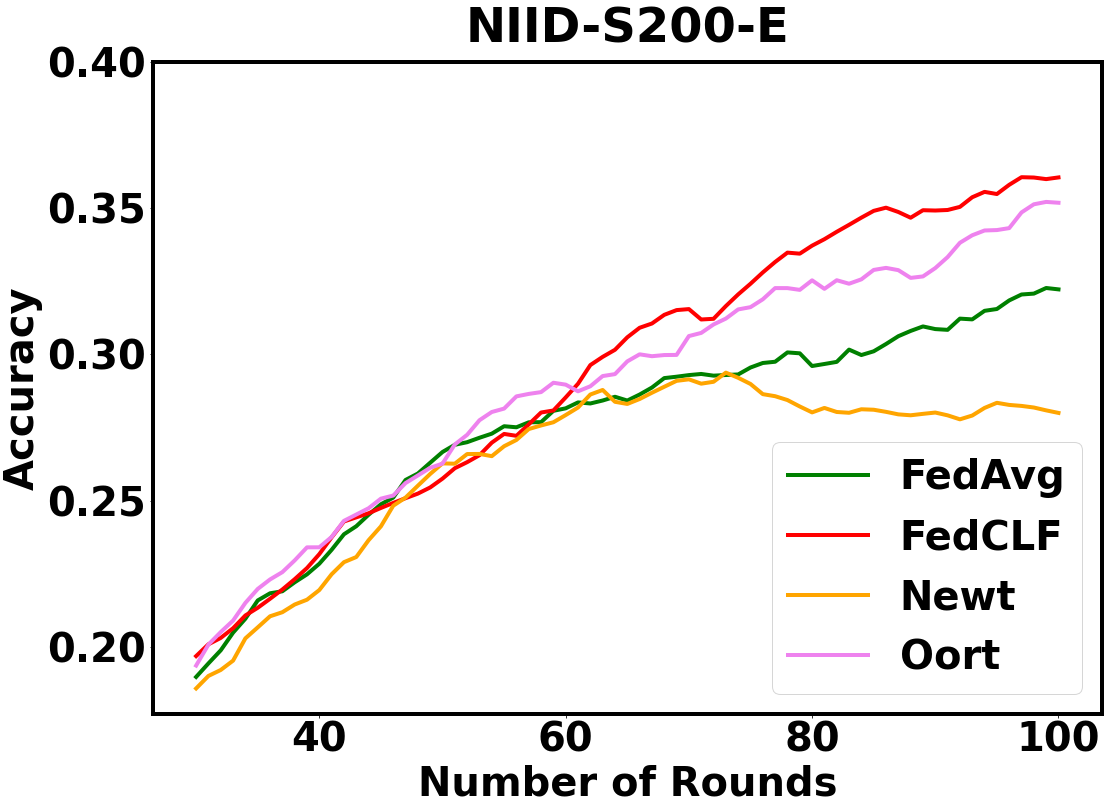}
         \caption{}
         \label{fig:g}
    \end{subfigure}
    \hfill
    \begin{subfigure}[b]{0.49\textwidth}
         \centering
         \includegraphics[width=\textwidth]{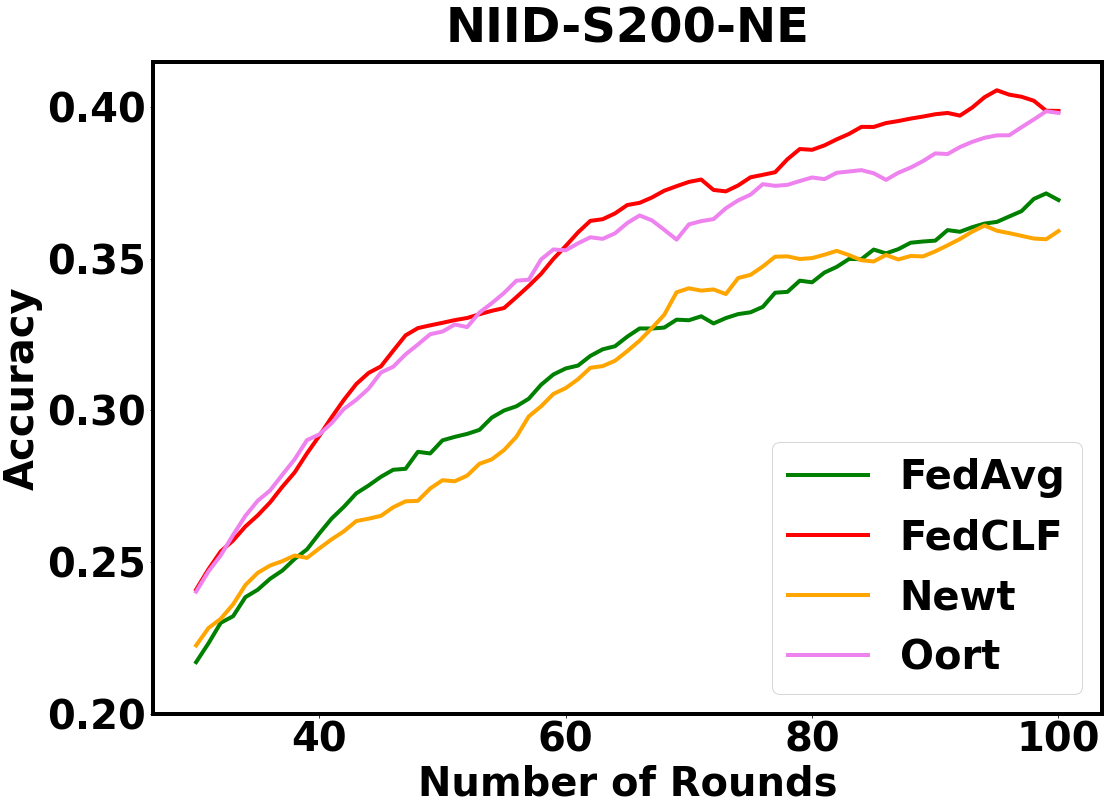}
         \caption{}
         \label{fig:h}
    \end{subfigure}
    \hfill
    
    \caption{Comparison of moving average accuracy of FedCLF with other state-of-the-art models for different types of datasets.}

    \label{fig:Overall Evaluation}
\end{figure}

In this section, we discuss how FedCLF outperforms other baseline models. Simulations were conducted on previously developed datasets, as depicted in Table \ref{tab:datasets}, to evaluate the performance of our method under different degrees of data heterogeneity. The final results are compared with the baseline models FedAvg, Oort and Newt, and are illustrated in Fig. \ref{fig:Overall Evaluation}. When the heterogeneity level of the dataset increases, the performance of FedCLF improves significantly over FedAvg, with gains ranging from approximately \textit{1\%} (IID-E) to \textit{16\%} (NIID-S200-E and NIID-S200-NE). Compared to Newt which was introduced in 2023 specifically targeting intelligent transport systems, FedCLF consistently demonstrates stable and significant improvements across all levels of data heterogeneity. Newt only shows notable performance on the NIID-S5-NE dataset, but even then, its performance drops significantly over time. Regarding Oort, it performs well as heterogeneity increases (NIID-S50-E/NE and NIID-S200-E/NE) but falls short compared to FedAvg when heterogeneity decreases (IID-NE and NIID-S5-NE). When comparing FedCLF to Oort, FedCLF consistently outperforms Oort, with improvements of around \textit{5\%} on datasets (NIID-S200-E and NIID-S200-NE) with higher level of data heterogeneity.

Another important advantage of employing FedCLF over the three baseline models considered is that FedCLF does not sample clients at each round like the baseline models. Instead, it samples only when the overall model accuracy degrades. This feature can be crucial, especially in resource-constrained environments, as it minimizes resource utilization while enhancing the efficiency of the FL training process. This advantage can be further verified quantitatively by plotting the overall model accuracy against the training time.

\section{Conclusion and Future Directions}\label{Conclusion}

The notion of Internet of Vehicles (IoV) plays a critical role in technological evolution and is expected to advance with applications like Federated Learning (FL). However, IoV faces several significant challenges, including a highly dynamic environment, high data and device heterogeneity, and the need of rapid model convergence due to safety and time critical applications. In this paper, we present FedCLF (FL with Calibrated Loss and Feedback control), an efficient participant selection process for FL tailored to the IoV networks. Our envisaged system introduces a novel utility measurement that calibrates loss to improve statistical efficiency in highly heterogeneous data environments. Additionally, we incorporate a feedback control mechanism that adjusts the sampling frequency based on changes in the overall model accuracy. This approach improves the efficiency of FL process as it optimizes the resource utilization while maintaining high accuracy. Further, this feedback control mechanism enhances the system's dynamism, allowing it to adapt to the surrounding environment's changes. Our results demonstrate that FedCLF significantly improves the overall model accuracy over state-of-the-art baseline models, showing remarkable performance in highly heterogeneous environments with up to a \textit{16\%} improvement. Moreover, we prove that the system's dynamic behavior, enabled by the feedback control mechanism, effectively reduces the sampling frequency while improving the overall model accuracy, leading to optimized resource utilization.

A limitation of our study is the use of a static dataset like CIFAR-10. Future work should evaluate FedCLF using dynamic datasets representative of IoV networks. Additionally, exploring dynamic threshold values instead of considering static thresholds, or comparing actual results with predicted values as feedback signals could yield further improvements.

\bibliography{mybibliography.bib}{}
\bibliographystyle{IEEEtran}

\end{document}